\documentclass[12pt]{article}

\usepackage{newtxtext,newtxmath}

\usepackage{graphicx}
\usepackage{caption}
\usepackage{tabularx}
\usepackage{booktabs}
\usepackage[letterpaper,margin=1in]{geometry}

\renewenvironment{abstract}
	{\quotation}
	{\endquotation}

\date{}

\makeatletter
\renewcommand{\fnum@figure}{\textbf{Figure \thefigure}}
\renewcommand{\fnum@table}{\textbf{Table \thetable}}
\makeatother

\usepackage{scicite}

\usepackage{url}

\DeclareCaptionFont{ninept}{\fontsize{10}{10}\selectfont}
\def\scititle{
    Grounding large language models as generalizable policies in  network control
}
\title{\bfseries \boldmath \scititle}

\author{
	Duo Wu$^{1,7}$,
	Linjia Kang$^{1}$,
	Zhimin Wang$^{1,7}$,
    Fangxin Wang$^{2}$,\and
    Wei Zhang$^{3}$,
    Chongbo Sun$^{1}$,
    Xuefeng Tao$^{3}$,
    Wei Yang$^{4}$,\and
    Le Zhang$^{5}$,
    Wenwu Zhu$^{6}$,
    Peng Cui$^{6}$,
    Zhi Wang$^{1\ast}$\and
	\small$^{1}$Shenzhen International Graduate School, Tsinghua University, Shenzhen \& 518055, China.\and
	\small$^{2}$School of Science and Engineering, The Chinese University of Hong Kong, Shenzhen, \and \small{Shenzhen \& 518172, China.}\and
	\small$^{3}$ByteDance, Shenzhen \& 518000, China.\and
	\small$^{4}$ByteDance, Hangzhou \& 311100, China.\and
	\small$^{5}$ByteDance, San Jose \& 95110, USA.\and
	\small$^{6}$Department of Computer Science and Technology, Tsinghua University, Beijing \& 100084, China.\and
	\small$^{7}$Pengcheng Laboratory, Shenzhen \& 518055, China.\and
	\small$^\ast$Corresponding author. Email: wangzhi@sz.tsinghua.edu.cn\and
}

\begin{document} 

\maketitle

\begin{abstract} \bfseries \boldmath

Designing generalizable control policies is essential for robust network services in modern digital infrastructure. Yet network control remains dominated by handcrafted rules or specialized deep learning models that struggle to generalize under real-world dynamics. Large language models (LLMs) offer a promising alternative because of their broad pretrained knowledge and emergent generalization abilities, but their practical adoption in network control is hindered by non-textual observations, constrained action spaces, complex optimization knowledge, and strict real-time requirements. Therefore, we introduce Trailblazer, a systematic framework that combines domain alignment to adapt LLMs for network control with adaptive policy collaboration to reduce inference overhead. Simulations across heterogeneous control tasks show that Trailblazer improves performance over conventional policies by 3.5\%-41.3\%. Moreover, in the production-grade evaluation on Douyin, Trailblazer outperforms a highly optimized industrial policy, corresponding to a projected reduction of 3,145 hours of platform-wide stall time per day. Further analysis reveals that effective LLM-based control can be achieved by properly aligning compact LLMs and only invoking them for complex conditions, rather than model scaling or extensive invocation. Together, our results establish an LLM-driven paradigm for designing generalizable network policies and offer insights for grounding foundation models in broader scientific and engineering domains.

\end{abstract}

\noindent

\section*{Introduction}

\begin{figure}[t]
    \centering
    \includegraphics[width=0.99\textwidth]{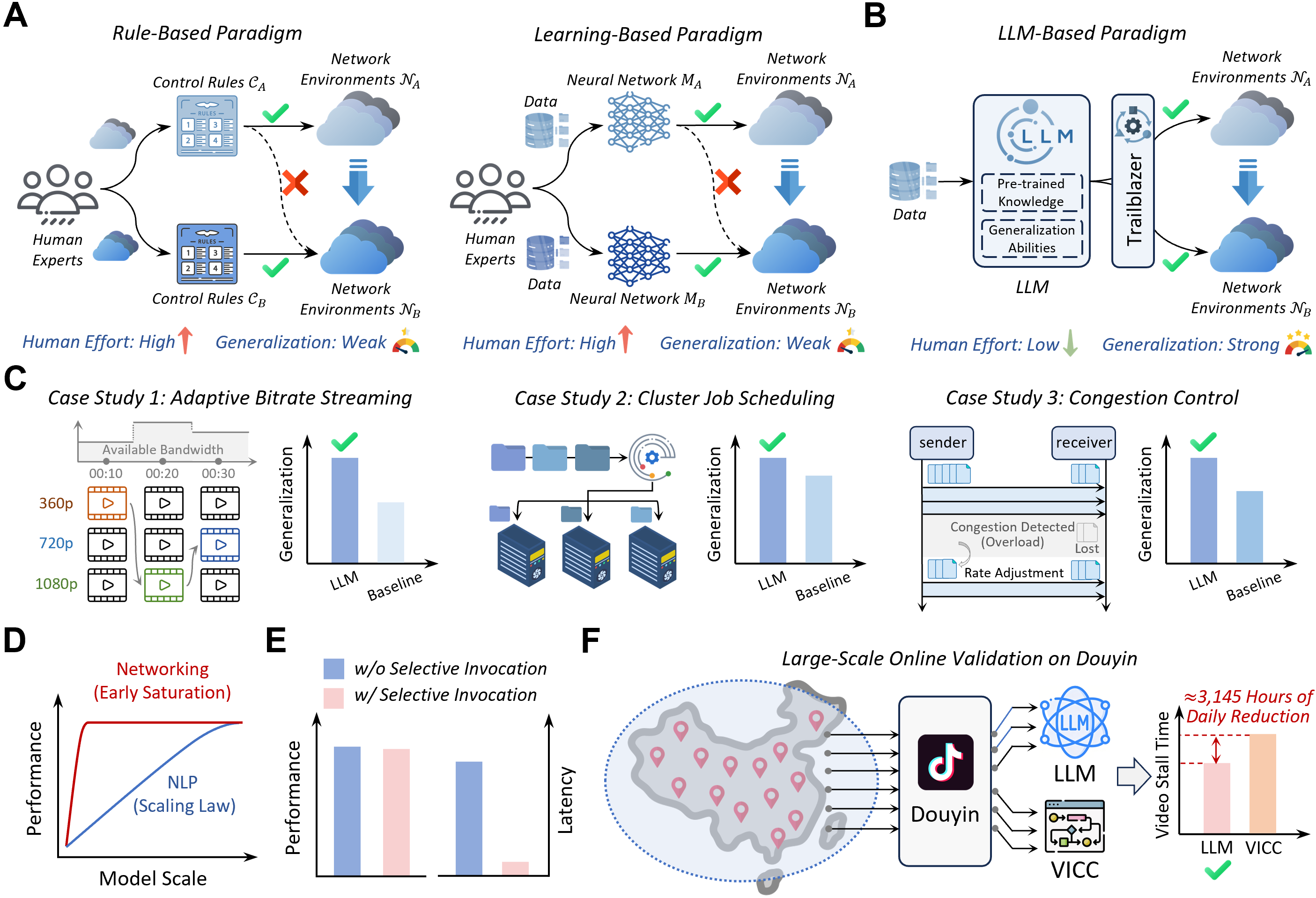}
    \vspace{-0.1cm}
    \caption{{\textbf{Motivation and contributions of this paper.} 
    (\textbf{A}) 
    Conventional paradigms for network policy design rely on either handcrafted rules or specialized learning models. Both necessitate manual updates or model retraining when network environments evolve beyond their original optimization scope, leading to high human effort and limited generalization.
    (\textbf{B}) 
    The proposed LLM-based paradigm. We design Trailblazer to ground LLMs as network policies. 
    By harnessing the pretrained knowledge and emergent generalization abilities of LLMs, Trailblazer enables robust performance across diverse environments, thereby achieving strong generalization with minimal engineering overhead for policy updates.
    (\textbf{C}) 
    Validation across three  heterogeneous networking tasks. Trailblazer enables LLMs to significantly outperform conventional rule-based and learning-based policies, demonstrating superior generalization. 
    (\textbf{D}) 
    Characterization of the early saturation phenomenon. In contrast to the scaling law in NLP, LLMs quickly reach a performance plateau in networking as the model scale increases, suggesting that compact LLMs can enhance inference efficiency without compromising performance. 
    (\textbf{E}) 
    Characterization of the selective invocation principle. Selectively invoking LLMs for complex scenarios while offloading simple cases to lightweight conventional policies can significantly reduce overall inference latency with negligible performance loss, reconciling model intelligence with strict latency constraints of real-world network control. 
    (\textbf{F}) 
    Large-scale online validation on Douyin. Building upon early saturation and selective invocation, we deploy the Trailblazer-enabled LLM in high-concurrency and latency-sensitive real-world environments.
    Compared to the mature industrial policy VICC, Trailblazer improves service quality under real-world dynamics, corresponding to an estimated daily reduction of 3,145 hours of video stall time.
    }}
    \label{fig:evoluation}
\end{figure}

Modern infrastructure and autonomous systems increasingly rely on decision-making mechanisms that must operate reliably under changing conditions, from chemical reactors and robotic platforms to smart grids and communication networks~\cite{bloor2025control,triantafyllidis2023hybrid,yu2025safe,ma2024efficient}. These mechanisms are often implemented as control policies that map observed states to actions, such as adjusting cooling-jacket temperatures, coordinating robotic motions, balancing electricity supply, or allocating network bandwidth. 
A central challenge is whether such policies can generalize beyond the intended conditions for which they were designed~\cite{huang2025foundation}. Computer networks are a compelling and practically important domain where this challenge arises at massive scale. As the backbone of modern digital infrastructure, network systems underpin global communications, large-scale data analytics, and various social platforms, requiring control policies to dynamically adjust system behaviors under rapidly changing conditions, such as adapting data transmission rates in response to bandwidth fluctuations.
However, as shown in Fig.~\ref{fig:evoluation}A, current paradigms for policy design typically rely on handcrafted control rules derived from human experts~\cite{li2014panda,venkat2018copa,yin2015mpc}, or deep learning models trained on specific network data~\cite{chen2018auto,mao2019cjs,kan2022improving,tan2024bwe}. The reliance on static priors or limited training data leads to a profound performance bottleneck: these policies are inherently coupled to the specific environments they are tuned for. Consequently, they often exhibit poor generalization when encountering the unpredictable dynamics of real-world networks, which frequently fall outside the scope of their original rules or data distributions~\cite{kan2022improving,huang2025chatNet}.
The limited generalization  necessitates continuous policy refinement in a trial-and-error manner, where  control rules are manually updated and  models are retrained, followed by extensive validation and iterative modification to accommodate the changes of network environments~\cite{mao2019cjs,chen2018auto,yen2023computers,meng2020interpreting,huang2025chatNet}.
This process requires substantial domain expertise  and  imposes  significant engineering overhead.
A concrete example is congestion control, the ``invisible hand'' that adjusts data transmission rates to avoid network overload and maintain stable Internet services~\cite{li2024congestion}. Despite decades of development, congestion control has undergone continuous refinement as Internet applications, traffic patterns, and deployment environments evolve, with the Internet Engineering Task Force (IETF) publishing 166 related technical documents since 2000~\cite{ietf2026cc}. This illustrates a long-standing  challenge in control policy design: static rules and narrowly trained models often require substantial human effort to remain effective under changing real-world conditions.
Consequently, the networking field remains constrained by a paradigm of constant manual adaptation and requires a transition toward network intelligence that can generalize across diverse real-world settings.

Motivated by the remarkable success of pretrained large models~\cite{lin2024vila,min2023recent}, we hypothesize that the key to overcoming the above limitations lies in a foundation model capable of encoding diverse network optimization principles. 
Large language models (LLMs)~\cite{grattafiori2024llama,qwen2025qwen25technicalreport,wei2021finetuned} offer a transformative path toward realizing this vision. Pretrained on Internet-scale corpora, including network textbooks and technical documents, LLMs implicitly encode general network knowledge within their parameters, forming a rich and unified knowledge base. Furthermore, they exhibit strong emergent abilities in pattern recognition and generalization to unseen scenarios~\cite{wei2022emergentabilities,xu2025reasoning,yang2024llmgeneralization} that have been validated across diverse scientific disciplines such as robotic control~\cite{kim2025openvla} and chemical synthesis~\cite{kim2024large}. These properties make it possible to leverage LLMs as the foundation for robust network policies that achieve strong generalization across diverse conditions. 

Although LLMs hold great promise for network optimization, their direct application faces two fundamental challenges. 
The first challenge lies in the \textit{domain misalignment} between LLMs and networking in input modalities, output validity, and optimization knowledge. 
Specifically, LLMs are primarily designed to process text, whereas network optimization relies on non-textual system observations, such as time-series bandwidth fluctuations. Meanwhile, their free-form token generation may fail to produce valid control actions within task-specific action spaces, raising reliability concerns for real-world network control. Moreover, while LLMs possess foundational network knowledge (e.g., concepts of bandwidth and congestion), this knowledge alone may be insufficient for complex tasks that require fine-grained control logic, such as determining appropriate transmission rates under congestion.
The second challenge stems from the \textit{computational inefficiency} of LLMs. If LLMs were invoked for every control decision (e.g., per-flow transmission rate adjustment), their high inference latency~\cite{Agrawal2024llmlatency,yu2022llmlatency,Kwon2023llmlatency} would render them impractical for deployment in real-time network services with strict latency constraints. 
While prior efforts have begun to explore the integration of LLMs into networking~\cite{liu2025llm4wm,jiao2025AI2MMUM,pokhrel2026distilling,wang2025bwe,yang2026cooperative,lin2025respond}, most are confined to task-specific designs such as network queue management~\cite{pokhrel2026distilling} and  traffic classification~\cite{lin2025respond}, providing few general insights in transforming LLMs into network policies for  broader networking contexts or other scientific disciplines. Besides, these studies, including the precursor of our work~\cite{wu2024netllm}, remain limited in simulation-level evaluations, leaving whether and how LLMs can be deployed for latency-sensitive real-world network control unexplored. 
Hence, this paper intends to comprehensively investigate the following key question: \textit{how can LLMs be effectively adapted as generalizable network policies  while satisfying the stringent latency requirements of real-world network control?}

\begin{figure}[t]
    \centering
    \includegraphics[width=0.99\textwidth]{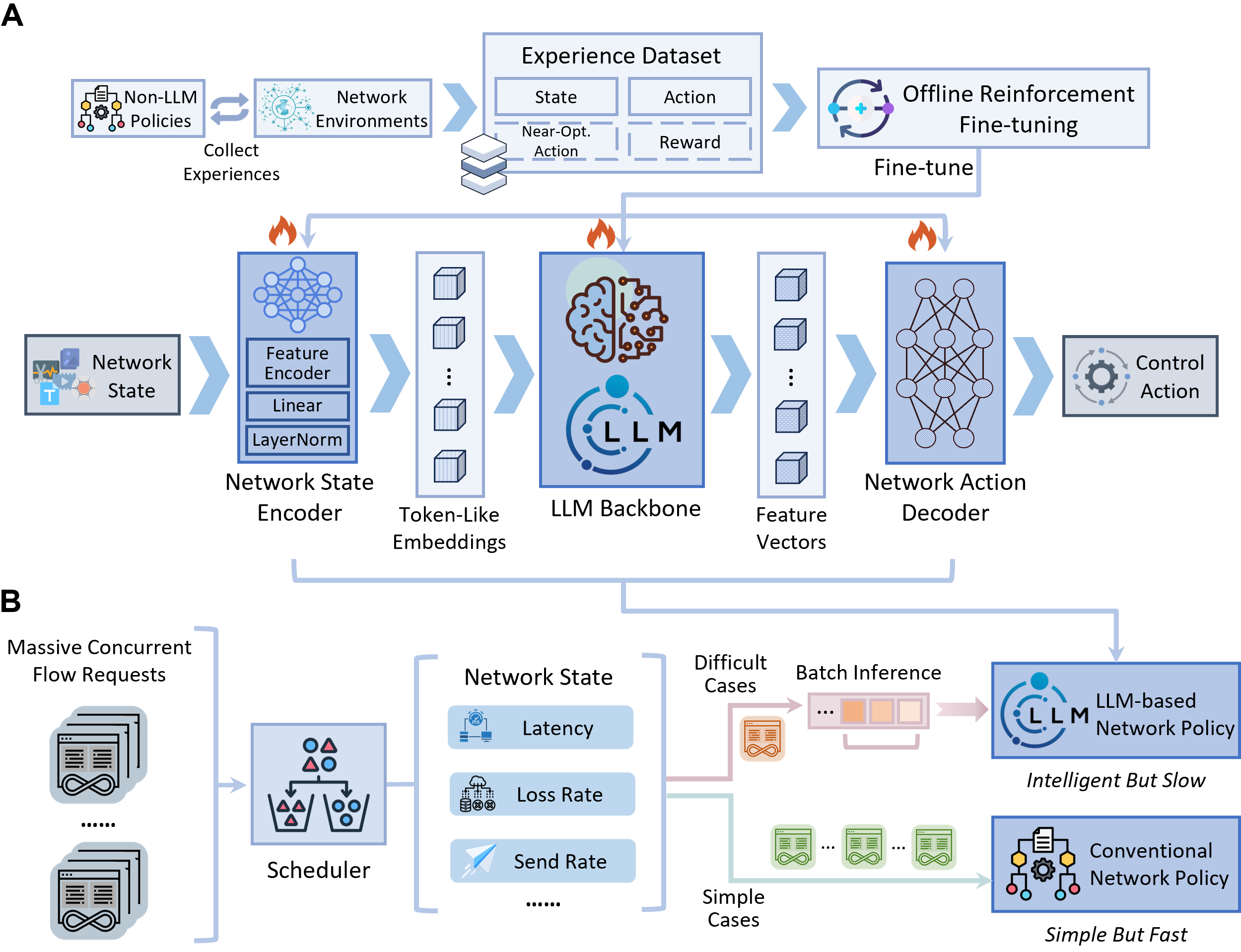}
    \caption{\textbf{Overview of the proposed Trailblazer framework.} 
    Trailblazer consists of two complementary modules to ground LLMs in network control: a domain alignment scheme for LLM adaptation and an adaptive policy collaboration mechanism for efficient deployment.
    (\textbf{A}) 
    The domain alignment scheme in Trailblazer, which includes a network state encoder to process multimodal network information, a network action decoder to generate specific control actions, as well as an offline reinforcement fine-tuning algorithm for the LLM to acquire domain knowledge.
    (\textbf{B})  
     The adaptive policy collaboration mechanism in Trailblazer, where the fine-tuned LLM collaborates with a lightweight conventional policy for intelligent and efficient network control. At its core, a scheduler dynamically routes flow requests based on real-time network conditions and offloads simple cases from the LLM to the lightweight policy for fast processing.
    }
    \label{fig:framework}
\end{figure}

This study overcomes the aforementioned barriers and establishes a general framework Trailblazer for the efficient adaptation and deployment of LLMs in network optimization, as shown in Fig.~\ref{fig:evoluation}B. To address the challenge of misalignment, Trailblazer incorporates a \textit{domain alignment scheme} to adapt LLMs for networking. 
It features a network state encoder to enable LLMs to process non-textual network observations analogously to linguistic tokens, and replaces free-form language generation with a network action decoder that directly produces valid, task-specific control decisions within predefined action spaces.
To inject domain-specific network knowledge into LLMs, Trailblazer further employs an offline reinforcement fine-tuning algorithm to fine-tune LLMs on an experience dataset comprising decision trajectories of diverse policies evaluated under various network environments. Guided by reward signals and expert-level actions, LLMs  learn to  discover high-performance control policies. To evaluate the effectiveness of Trailblazer in LLM adaptation, we conduct extensive experiments under diverse and realistic conditions on two representative networking tasks of social and industrial relevance: adaptive bitrate streaming (ABR)~\cite{timmerer2025http} for optimizing video services and cluster job scheduling (CJS) supporting large-scale data analytics~\cite{sanjalawe2025ai} (see Fig.~\ref{fig:evoluation}C). Experimental results demonstrate that the Trailblazer-adapted LLM successfully outperforms existing state-of-the-art methods on various network environments by 6.5\%-36.6\% in ABR and 3.5\%-41.3\% in CJS, demonstrating stronger generalization. Building on these outcomes, we extend our investigation beyond functional validation to explore the underlying scaling and deployment characteristics of LLMs in production-grade network services. We select congestion control (CC)~\cite{li2024congestion} as a representative case study due to its significant importance for modern digital services and stringent real-time requirements. As depicted in Fig.~\ref{fig:evoluation}D, we reveal an interesting phenomenon, \textit{early saturation}, which suggests that compact LLMs can be effectively leveraged for network control to significantly enhance inference efficiency without compromising performance. Furthermore, we uncover another fundamental principle for practical LLM deployment--\textit{selective invocation} (see Fig.~\ref{fig:evoluation}E), i.e., invoking the LLM for network control only under complex conditions, rather than on every request. These findings jointly reveal that the key to LLM-driven network control is not simply scaling models up or invoking them for every request. Instead, once properly aligned, compact LLMs can be positioned as intelligent experts selectively invoked for complex network conditions. 

We translate these scientific insights into the design of \textit{adaptive policy collaboration mechanism} in Trailblazer for efficient LLM deployment. The core idea  is a scheduler to enable efficient collaboration between a compact LLM and a conventional rule-based policy. The scheduler evaluates the network conditions of each incoming request and selectively routes those under poor conditions to the LLM for intelligent control, while directing the rest to a lightweight rule-based policy for fast processing. To validate this paradigm at scale, we adapted and deployed the LLM with Trailblazer within the CC service in Douyin (the Chinese version of TikTok). We conducted large-scale online A/B tests to compare the LLM against a strong and mature industrial policy VICC. Over the three-week evaluation period, both systems operated in parallel. Each served over 150,000 users, spanning more than 100 cities and collectively accumulating over 28,800 hours of video playback time. Results showed that, with Trailblazer, the LLM consistently outperformed VICC across all industrial key performance metrics, which can be translated to a  daily reduction of 3,145 hours of video stall time across the platform, as illustrated in Fig.~\ref{fig:evoluation}F. This evaluation provides production-scale evidence that a domain-aligned LLM can operate within a real-world control loop and deliver measurable benefits over the industrial policy refined through years of deployment.

Taken together, our findings establish LLMs as a promising foundation for next-generation network policies, paving the way for a shift from conventional rule-based or learning-based paradigms toward an LLM-driven paradigm for generalizable network control.
Importantly, the large-scale deployment on Douyin provides production-grade evidence that such a paradigm can operate under real-world concurrency, latency, and reliability constraints.
Furthermore, our work provides a practical blueprint for the efficient adaptation and  deployment of LLMs in real-world control systems, uncovering a set of design principles for domain alignment, model scale selection and efficient execution. These findings suggest a pathway for grounding foundation models as adaptive control policies in dynamic environments, with potential implications for broader scientific and engineering  domains such as chemical process control, robotic manipulation and power grid management, where generalization, responsiveness, and operational reliability are essential. 


\section*{Results}
In this section, we first introduce Trailblazer, the proposed framework for adapting and deploying LLMs in network control. We then validate its ability to support generalizable network control in simulated environments, where heterogeneous tasks and distribution shifts can be systematically evaluated. Next, we analyze the mechanisms underlying this performance, including the roles of pretrained knowledge, domain-specific adaptation, model family, and model scale. This analysis motivates two deployment principles, early saturation and selective invocation, which guide the final large-scale online evaluation on Douyin.

\subsection*{Trailblazer framework}
The misalignment between LLMs and networking and their high inference latency pose significant challenges to the adaptation and deployment of LLMs in networking. To systematically address these challenges, the proposed Trailblazer framework (Fig.~\ref{fig:framework}) consists of two complementary modules: a domain alignment scheme to adapt LLMs for network control and an adaptive policy collaboration mechanism for  efficient deployment. 

Fig.~\ref{fig:framework}A details the domain alignment scheme in Trailblazer, which transforms the LLM into a network control policy through three core components: a network state encoder, a network action decoder, and an offline reinforcement fine-tuning algorithm. 
To bridge the input modality gap, a network state encoder extracts features from raw network statistics of various modalities using modality-specific feature encoders, and maps the extracted features into the same semantic feature space as language tokens with a linear projection layer. Subsequently, layer normalization is used to fuse different features, producing a set of token-like embedding vectors. This enables the LLM to interpret non-linguistic network information as if processing natural language. 
To address output misalignment, Trailblazer replaces the LLM’s original prediction head for language tokens with a dedicated network action decoder, which transforms the LLM-produced high-dimensional feature vectors into valid network control actions within predefined task-specific action spaces. This constrained action-generation design structurally mitigates the risk of hallucinated or physically invalid control decisions.
Both the state encoder and action decoder are trainable to learn optimal projection functions. With the above adaptation, the LLM is able to process network states and generate network actions.  Fig.~\ref{fig:figures1} provides a concrete task-specific adaptation example. As for knowledge alignment, Trailblazer fine-tunes the LLM to acquire domain knowledge for networking with an offline reinforcement fine-tuning algorithm. Specifically, we evaluate conventional non-LLM network policies across various network environments and collect their interactions with the environments, i.e., network state-action pairs and the associated rewards or near-optimal actions, as the experience dataset. This dataset serves as the training fuel for fine-tuning the LLM. By using rewards or near-optimal actions as guiding signals, the LLM can analyze the behavior of existing policies, learn from effective actions, and uncover the reasons behind poor actions, thereby automatically discovering a better-performing policy. 

Fig.~\ref{fig:framework}B depicts the adaptive policy collaboration mechanism in Trailblazer, which establishes a hybrid control paradigm to reconcile the high intelligence of LLMs with strict real-time networking constraints. The key idea of this module is to use a scheduler to selectively offload flow requests from the LLM  to a lightweight rule-based policy for fast processing, thereby accelerating the processing speed of the LLM-based network systems. Specifically, for each incoming request, the scheduler evaluates its network conditions and only routes those under poor network conditions to the LLM. The rest under stable conditions are instead handled efficiently by a conventional policy. Under this mechanism, the collaboration between the LLM and conventional policy contributes to a more robust network system that exploits the strong capabilities of the LLM to address complex scenarios while offloading a large volume of simple requests to a lightweight policy to improve the overall efficiency. To further improve the system efficiency, the LLM is typically of relatively small scale (e.g., 0.5B parameter size) and processes requests in batches to reduce the per-request processing latency.

\subsection*{Simulation-based validation of LLM adaptation}

\begin{figure}[t]
    \centering
    \includegraphics[width=1.0\textwidth]{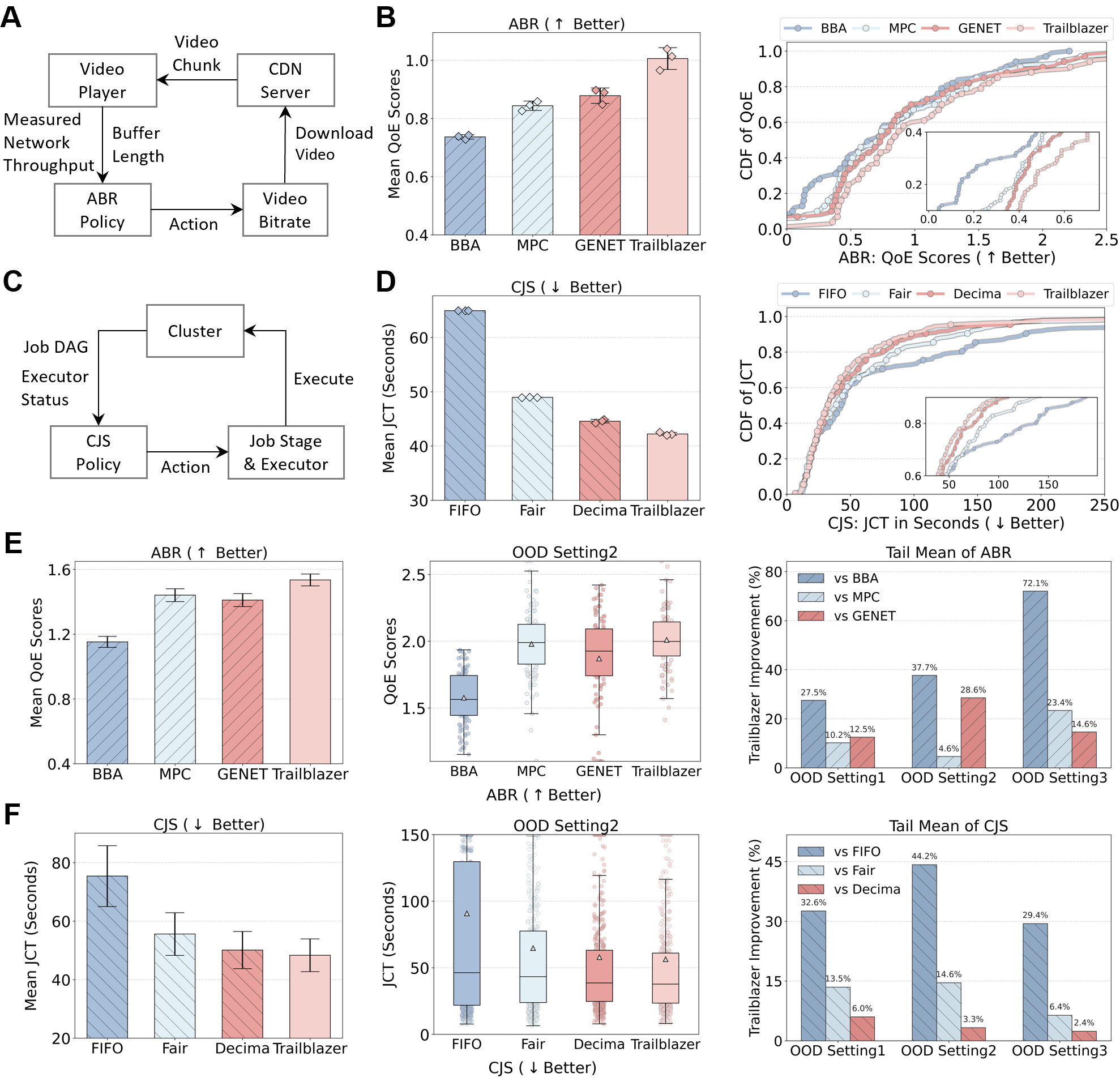}
    \vspace{-0.9cm}
    \caption{\textbf{{Comprehensive generalization comparison between the Trailblazer-adapted LLM  and baselines in heterogeneous networking tasks and environments.}}  
    (\textbf{A} and \textbf{C}) 
    Schematic illustrations of the ABR and CJS tasks, respectively. 
    (\textbf{B} and \textbf{D}) 
    Evaluation under ID test settings matching the training configurations. 
    The left panels report the mean performance of each method averaged over three random seeds. The right panels visualize the distribution of performance metrics. 
    (\textbf{E} and \textbf{F})
    Evaluation  under various  more challenging OOD test settings. 
    The left panels report the mean performance across three OOD settings. The middle panels show the performance distribution of each method on OOD setting 2, where triangles denote mean values. The right panels report the performance improvement of Trailblazer over baselines in terms of tail mean (the worst 25\% cases) under different OOD settings.
    }
    \label{fig:general_evaluation}
\end{figure}

In this section, we systematically evaluate the effectiveness of Trailblazer in LLM adaptation. As shown in Fig.~\ref{fig:general_evaluation}A and~\ref{fig:general_evaluation}C, two representative tasks with social and industrial relevance are selected for evaluation: ABR and CJS. ABR~\cite{xia2022genet} is a cornerstone of online video platforms like TikTok, optimizing video bitrates under dynamic network conditions and playback status. CJS~\cite{mao2019cjs} is  crucial for high-performance computing, efficiently distributing computational workloads across multiple cluster nodes to support large-scale data analytics. These tasks cover the distributed and centralized control paradigms in networking: the ABR clients independently optimize bitrates while the CJS controller is responsible for the entire cluster. Besides, as illustrated in Table S1, they are highly heterogeneous in terms of input modalities, action spaces, and optimization objectives. Hence, these characteristics make our subsequent findings representative and applicable to a wide range of scenarios. Next, considering that simulation offers a flexible and controllable approach to emulate diverse conditions for systematic evaluation, we simulate various network environments spanning both in-distribution (ID) and out-of-distribution (OOD) environments based on real-world datasets for comprehensive generalization evaluation. By default, we use Llama2-7B~\cite{touvron2023llama2},  the state-of-the-art LLM at the time of this evaluation, as the foundational LLM of Trailblazer. In each task, we compare the Trailblazer-adapted LLM with one  learning-based policy and two rule-based policies: GENET~\cite{xia2022genet}, BBA~\cite{huang2014bba}, MPC~\cite{yin2015mpc} for ABR and Decima~\cite{mao2019cjs}, FIFO~\cite{spark_scheduler}, Fair~\cite{spark_scheduler} for CJS. These baselines are selected for their strong performance and open-source reproducibility. For performance metrics, quality of experience (QoE)~\cite{mao2017neural} that quantifies the user’s video watching experience is adopted for ABR, while job completion time (JCT)~\cite{mao2019cjs} is used to measure the duration of job execution with specific computing resources for CJS. Higher QoE and lower JCT indicate better performance. The details of setup, Trailblazer implementation, datasets, baselines and metrics are provided in the Methods section.

\subsubsection*{Generalization across diverse environments}


Fig.~\ref{fig:general_evaluation}B and~\ref{fig:general_evaluation}D show the performance of each method in ID test environments that share similar statistical distributions (e.g., bandwidth fluctuation patterns) to those in the training phase. The Trailblazer-adapted LLM consistently outperforms all baselines, achieving 14.5\%-36.6\% higher QoE in ABR and reducing JCT by 6.8\%-41.3\% in CJS. Results further reveal that the LLM achieves superior performance across the entire distribution while exhibiting strong resilience in tail performance. For example, compared to the learning-based policies, the LLM improves the 10th percentile QoE by 13.4\% in ABR and reduces the 90th percentile JCT by 11.0\% in CJS. Notably, while conventional methods rely on labor-intensive, task-specific designs such as handcrafted rules or specialized neural architectures, Trailblazer enables the optimization of heterogeneous tasks using a unified LLM through lightweight adaptation (i.e., state encoders and action decoders) and efficient offline fine-tuning. These results not only validate the effectiveness of Trailblazer in adapting LLMs for network optimization but also demonstrate the feasibility of using LLMs as the foundation for network control.

To evaluate generalization under distributional shifts, Fig.~\ref{fig:general_evaluation}E and~\ref{fig:general_evaluation}F present the performance of each method across multiple challenging OOD test environments that differ from training distributions. On average, the Trailblazer-adapted LLM consistently outperforms all baselines (left panels), improving mean QoE by 6.5\%-33.2\% in ABR and reducing mean JCT by 3.5\%-35.9\% in CJS. This superiority is further reflected in the overall performance distributions (middle panels for OOD Setting 2; full results in Fig.~\ref{fig:figures2}). In particular, results indicate that conventional learning-based policies do not consistently outperform rule-based approaches under OOD conditions. For instance, in the ABR task, GENET underperforms the rule-based MPC by 5.9\% in average QoE and exhibits a worse performance distribution on OOD Setting 2. This degradation highlights the vulnerability of standard deep learning models to unseen environments. In contrast, Trailblazer harnesses the rich pretrained knowledge of the foundational LLM to maintain robust generalization, achieving the highest QoE across all OOD settings. To explicitly assess worst-case reliability  of each method  under unpredictable network dynamics, we further analyze the tail performance, defined as the mean performance within the worst 25\% of cases. As shown in the  right panels of Fig.~\ref{fig:general_evaluation}E and~\ref{fig:general_evaluation}F, the Trailblazer-adapted LLM exhibits a substantially more robust lower bound. In ABR, it enhances the tail mean QoE by 4.6\%-72.1\% against rule-based policies and maintains a 12.5\%-28.6\% advantage over GENET across varying OOD settings. Similarly, in CJS, it reduces  the tail mean JCT by 6.4\%-44.2\% over rule-based policies and by 2.4\%-6.0\% over Decima. These results validate the superior generalization of the LLM-driven paradigm, demonstrating its capability to maintain  robust control and prevent performance collapse even in the most  unpredictable network environments--a critical requirement for real-world network services~\cite{yen2023computers}.

\begin{figure}[t]
    \centering
    \includegraphics[width=0.9\textwidth]{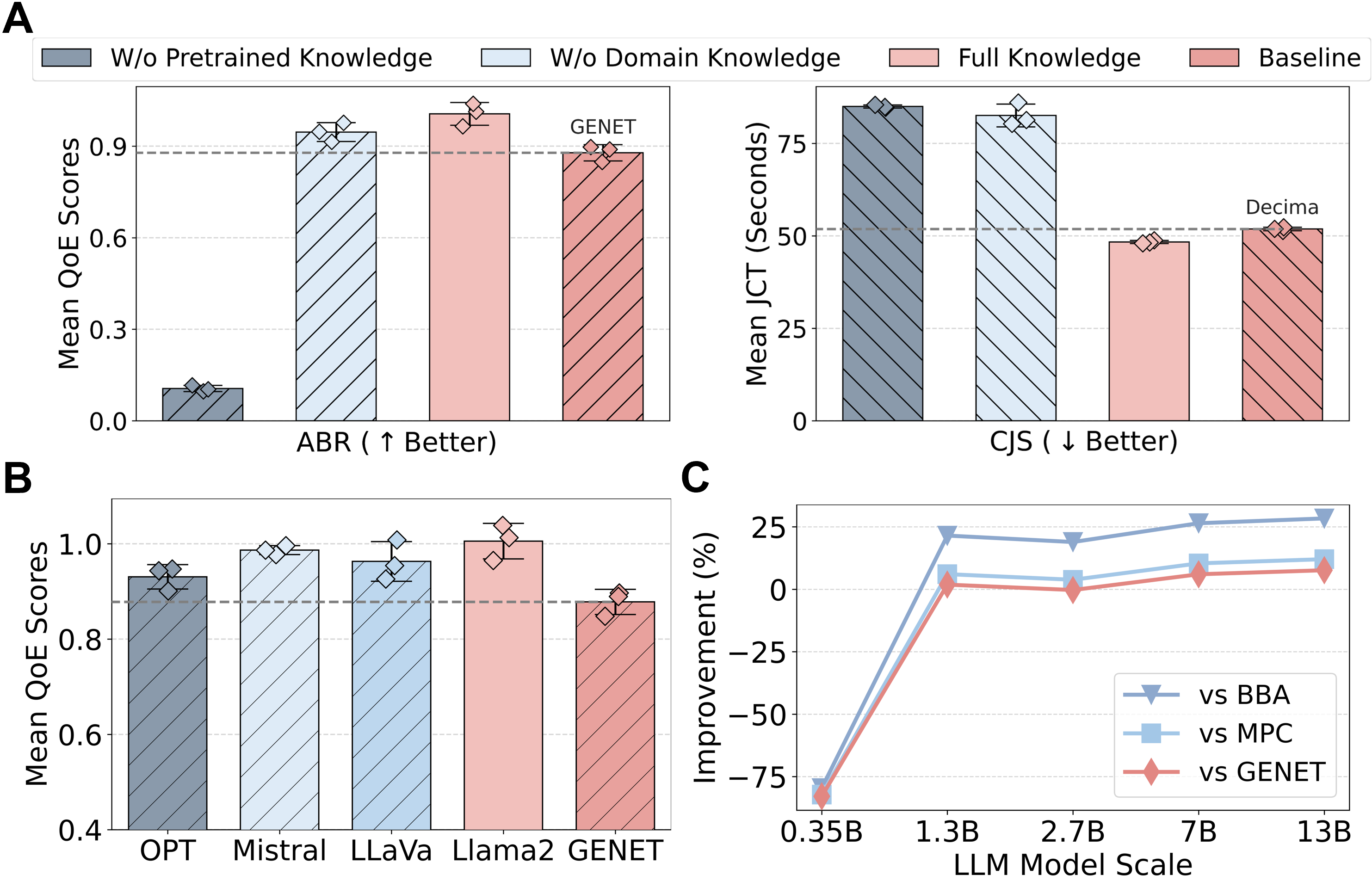}
    \vspace{-0.1cm}
    \caption{\textbf{Mechanistic analysis of LLM adaptation in networking.}
    (\textbf{A}) Investigation on the importance of pretrained knowledge of the LLM and domain knowledge injected by Trailblazer. 
    (\textbf{B}) Investigation on the performance of different LLM model families.
    (\textbf{C}) Investigation on the impact of LLM model scale on task performance.
}
    \label{fig:deep_dive}
\end{figure}

\subsubsection*{Mechanistic analysis of LLM adaptation}

To gain deeper insights into the effectiveness of LLMs in networking, we conduct a comprehensive ablation study focusing on three dimensions: knowledge source, model family, and parameter scale. First, to investigate the influence of pretrained and domain-specific knowledge in adapting LLMs, we construct two variants of Trailblazer. One discards the pretrained weights of the LLM that represent its pretrained knowledge by randomly reinitializing and training the LLM from scratch. The other preserves the pretrained weights but freezes the entire LLM backbone to prevent the direct acquisition of domain knowledge, only fine-tuning the network state encoder and action decoder. Second, to explore whether the previously observed performance advantages are specific to a particular family, we use the ABR task as an example and evaluate four representative open-source foundation models: Llama2~\cite{touvron2023llama2}, OPT~\cite{zhang2022opt}, Mistral~\cite{jiang2023mistral}, and LlaVa~\cite{liu2023visual}. All models are standardized to a 7B parameter scale to ensure a fair and consistent comparison. Finally, we investigate the impact of model scale on task performance by evaluating the OPT model family of varying parameter sizes.

Fig.~\ref{fig:deep_dive}A depicts the performance of the two variants without pretrained or domain knowledge in ABR and CJS tasks. The variant without pretrained knowledge suffers from significant performance degradation, validating that the pretrained knowledge of the LLM indeed contains common knowledge useful for networking and serves as a critical prerequisite for the LLM to function effectively in networking. Moreover, even when general pretrained knowledge is retained, the variant without domain knowledge still exhibits a noticeable performance drop. This highlights the critical importance of integrating domain-specific knowledge into the LLM for network optimization. Trailblazer plays a key role in bridging the gap between general knowledge and domain expertise. By jointly integrating these two facets, LLMs can be efficiently adapted to diverse networking tasks. Next, Fig.~\ref{fig:deep_dive}B presents the performance of different LLM families adapted via our Trailblazer framework. Results show that all fine-tuned LLMs significantly outperform the learning-based policy GENET. This indicates that the effectiveness of LLMs in networking is not limited to a specific model family and validates Trailblazer as a model-agnostic framework for LLM adaptation. Finally, Fig.~\ref{fig:deep_dive}C depicts the performance of the OPT models of varying parameter sizes in the ABR task. OPT-0.35B performs worse than baselines, potentially due to insufficient general knowledge. In contrast, all OPT variants larger than 1B outperform all baselines, but performance saturates rapidly beyond this threshold, with larger models yielding only marginal gains. This preliminary finding suggests that the capacity of the LLM required for networking may be relatively small. Thus, small LLMs can be utilized to achieve competitive performance with faster inference speed. This phenomenon is further systematically validated and characterized in the subsequent section.

\subsection*{From simulation to production deployment}



To assess whether the advantages of LLMs can transfer from simulation to real-world production systems, we integrate Trailblazer into the real-time CC service within Douyin, one of the world’s most popular video social platforms serving  tens of millions of daily active users.  CC is a fundamental networking task underpinning a wide range of modern applications, such as real-time video streaming and video calls. In network systems, data transmission can be analogized to water flowing through a series of pipes: end-to-end throughput is constrained by the slowest segment--the bottleneck link~\cite{kurose2021topdown}. CC aims to infer this typically unobservable link bandwidth using recent network measurements, enabling adaptive rate control to proactively avoid congestion and improve resource utilization~\cite{tan2024bwe,mei2020bwe,wang2025bwe}. In particular, real-world CC in Douyin requires end-to-end response latency below 100\,ms, making it significantly challenging for LLM deployment. Hence, before deploying the LLM in production-grade environments, we first systematically characterize two phenomena: early saturation and selective invocation, which are crucial for improving the inference efficiency of LLM-based network systems. Guided by these insights, we then conduct large-scale online tests to validate the generalization capabilities of the LLM. For the subsequent evaluation, we employ Qwen2.5~\cite{qwen2025qwen25technicalreport} as the default foundation LLM. We benchmark the Trailblazer-adapted LLM against VICC, a mature CC policy that has been deployed in production at Douyin and iteratively refined over several years. This comparison allows us to benchmark the Trailblazer-adapted LLM against a strong, real-world, industrial baseline. The details of implementation and experimental setup are provided in the Methods section.

\subsubsection*{Deployment principles: early saturation and selective invocation}
\begin{figure}[t]
    \centering
    \includegraphics[width=0.92\textwidth]{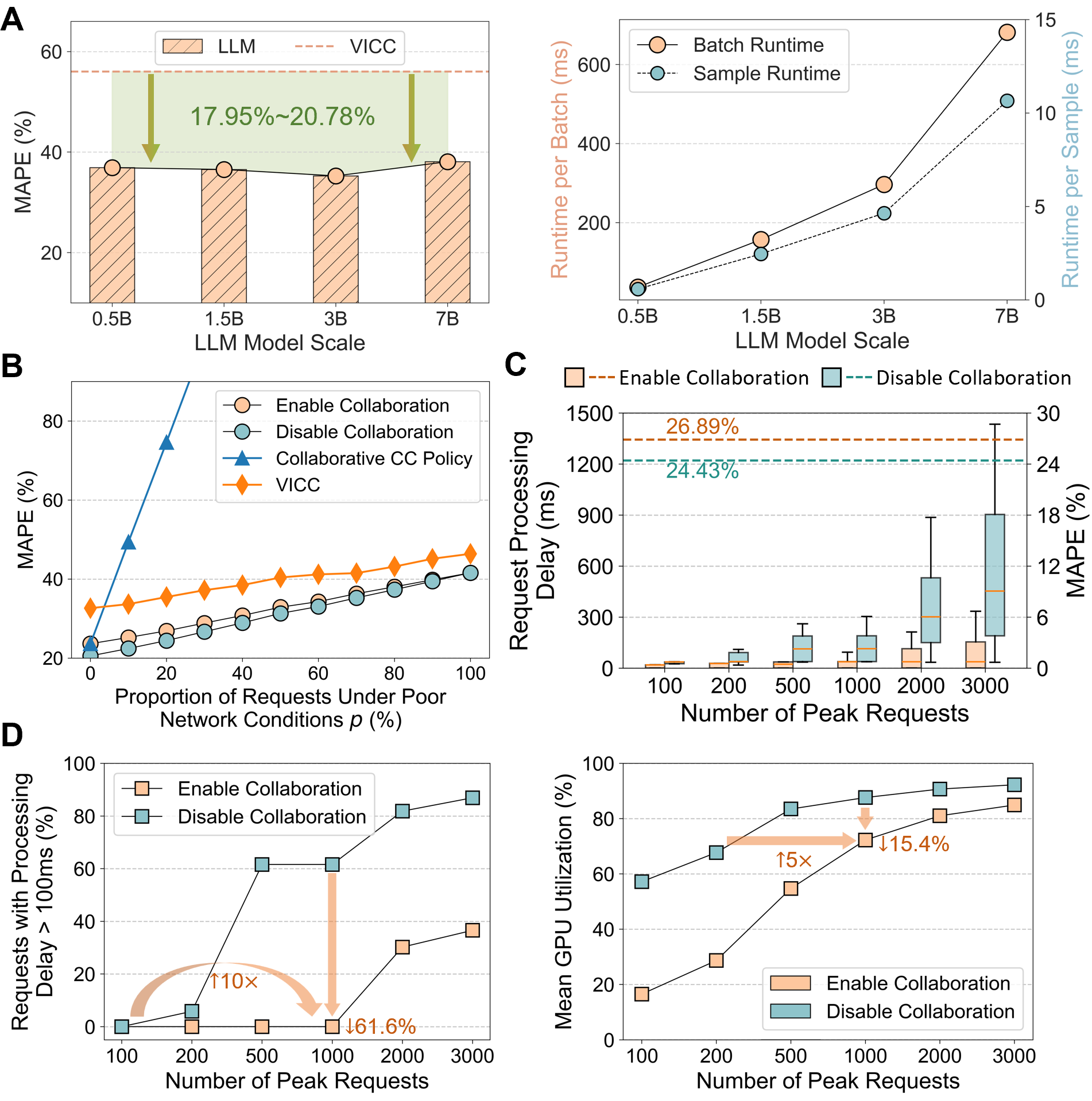}
    \vspace{-0.1cm}
    \caption{\textbf{Systematic characterization of the influence of model scales and policy collaboration.} 
    (\textbf{A}) Performance comparison across Trailblazer-adapted LLMs with varying model scales. 
    (\textbf{B}) Comparison between Trailblazer with and without collaboration, evaluated across different proportions ($p$) of requests under poor network conditions. 
    (\textbf{C}) Comparison between the two variants in terms of processing delay and task performance under varying peak request volumes when $p=20\%$. 
    (\textbf{D}) Comparison between the two variants in terms of fraction of requests with large delays and GPU utilization under varying peak request volumes when $p=20\%$.
    }
    \label{fig:insight}
\end{figure}

Motivated by the preliminary observations  in Fig.~\ref{fig:deep_dive}C, which suggest a performance plateau in the OPT family, we further investigate the impact of LLM model scale on the performance and computational cost in the CC task. Specifically, we compare Qwen2.5 with 0.5B, 1.5B, 3B, and 7B parameters against the baseline VICC.  The inference batch size is fixed as 64.  Performance is measured with mean absolute percentage error (MAPE) as the primary metric, which quantifies the relative discrepancy between the sending rate and bottleneck link capacity or request bandwidth. Lower MAPE indicates better performance. As shown in Fig. 5A, all variants outperform VICC, with the absolute MAPE reduction of 17.95\%-20.78\%. Notably, their performance quickly saturates with increasing model scales: starting from 0.5B parameters, MAPE remains around 36.5\% across all model scales. In contrast, the computational cost rises sharply with model size, with per-batch runtime increasing from 37 ms (0.5B) to 682 ms (7B)--an 18.4$\times$ increase. We formally term this phenomenon \textit{early saturation}, which reveals a general insight that the performance of LLM in networking reaches a stable plateau at a relatively small model scale, as evidenced by both OPT for ABR (Fig.~\ref{fig:deep_dive}C) and Qwen2.5 for CC (Fig.~\ref{fig:insight}A). 
This diverges from the typical scaling trends observed in NLP, where performance generally improves with model scale.
Early saturation indicates that the model scale of LLM required for effective network optimization is relatively small, and thus compact LLMs can be used for networking to improve efficiency without compromising performance. Consequently, based on this insight, we adopt the 0.5B Qwen2.5 model for all subsequent experiments to achieve the optimal inference efficiency.

Although exploiting early saturation significantly reduces the inference latency of individual requests, executing even a compact LLM on a per-request basis remains computationally prohibitive under massive production workloads.
To improve the operational efficiency of LLM-based network systems, one possible strategy is to synergize high-precision control with low-latency execution. Therefore, we investigate a collaborative framework where the Trailblazer-adapted LLM is complemented by a lightweight rule-based policy that can reliably resolve common cases in real-time. Specifically, we design a scheduler to enable collaboration, which adaptively offloads flow requests from the LLM to a rule-based policy according to real-time network states (e.g., whether packet loss remains below a predefined threshold). The details of the scheduler and lightweight policy are provided in the Methods section.

Fig.~\ref{fig:insight}B illustrates the MAPE of the Trailblazer-adapted LLM with or without the scheduler to enable collaboration, as the proportion of requests under poor network conditions ($p$) increases. As shown in Fig.~\ref{fig:insight}B, even when collaborating with a simple CC policy whose MAPE grows rapidly when $p$ increases, Trailblazer still significantly outperforms VICC across all cases. Besides, enabling collaboration only incurs at most 3.08\% higher MAPE than the variant without collaboration, and the performance gap narrows at higher $p$, as more requests are routed to the LLM. Fig.~\ref{fig:insight}C further compares the two systems in terms of request processing delay when $p = 20$\%. As request peak volume increases, Trailblazer without collaboration exhibits severe performance degradation with significantly higher processing delay. This is because without offloading requests to the lightweight policy, a large volume of requests are all directed to the LLM simultaneously, leading to long queuing delay. In contrast, Trailblazer with collaboration demonstrates strong resilience and superior efficiency. For instance, at $p = 20$\% and a peak load of 2,000 requests, Trailblazer with collaboration reduces the average delay from 345\,ms to 61\,ms--below Douyin’s CC requirement of 100\,ms--while incurring only a negligible 2.46\% increase in MAPE compared to Trailblazer without collaboration. 
Fig.~\ref{fig:insight}D further validates the efficiency of collaboration, where the number of requests processed under 100\,ms increases from 100 to 1000, representing a 10-fold improvement. A similar improvement is observed in GPU efficiency: at around 70\% utilization, the system can support 1000 peak requests, compared to only 200 when collaboration is disabled. 
These results reveal an important principle: invoking the LLM for network control when necessary rather than implementing per-request control. We term it \textit{selective invocation}, the key to improving the efficiency of LLM-based network policies in real-world network systems.

Taken together, these insights establish a practical paradigm for LLM deployment in networking. Early saturation justifies the use of compact models to optimize efficiency, and selective invocation shifts the inference paradigm from per-request execution to a hybrid, collaborative one. We complement the design of Trailblazer by formalizing these principles into the mechanism of adaptive policy collaboration, providing a general blueprint to harness the intelligence of LLMs while maintaining high operational efficiency in real-world network  systems. In the following section, we leverage this mechanism to deploy and validate the adapted LLM in a large-scale production environment.

\subsubsection*{Large-scale online evaluation on Douyin}

\begin{figure}[t]
    \centering
    \includegraphics[width=0.98\textwidth]{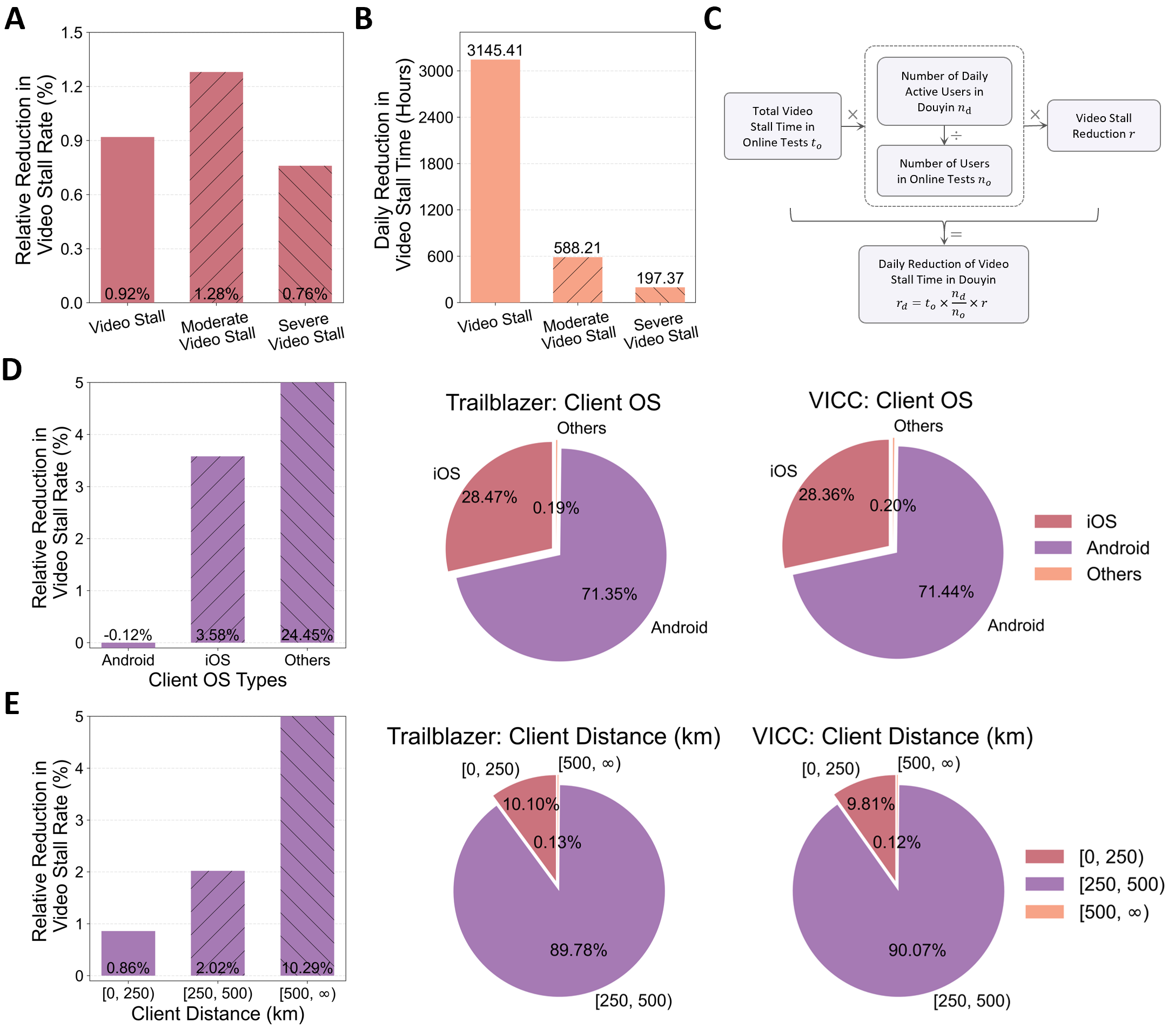}
    \vspace{-0.1cm}
    \caption{{\textbf{Results of large-scale online A/B tests between  Trailblazer and  VICC within Douyin's CC service.}}  
    (\textbf{A}) 
    Relative reduction of Trailblazer over VICC on different video stall rates. 
    (\textbf{B}) 
    The projected platform-wide daily reduction in video stall time.
    (\textbf{C})
    Mathematical framework used to estimate platform-wide stall time reduction. 
    (\textbf{D}) 
    Performance comparison across client operating systems. 
    (\textbf{E}) 
    Performance comparison across client-to-server geographical distances.
}
    \label{fig:online_test}
\end{figure}


We deployed the adapted LLM with  Trailblazer's adaptive policy collaboration mechanism  enabled in the online CC service of Douyin, which underpins its social video call application. The system operated in the online production environment for 3 weeks, during which  we conducted large-scale A/B tests to benchmark Trailblazer against VICC. The primary performance metric was video stall rate, a key industrial metric that directly impacts playback smoothness and user experience. Lower stall rates generally indicate superior CC performance. This is because by proactively and accurately estimating the available bottleneck link bandwidth, the application system can dynamically adjust the data transmission rate in response to network changes, thereby avoiding bandwidth overuse and preventing congestion-induced playback disruptions. We use this application-specific metric to investigate whether the LLM can introduce industrial benefits. In the following analysis, due to compliance with Douyin’s data security policy, all results are reported as relative reductions in video stall rates achieved by Trailblazer compared to VICC, without disclosing absolute metrics. The details of online deployment and performance metrics are provided in the Methods section.


Fig.~\ref{fig:online_test}A presents the results of our large-scale online tests in terms of video stall rate reduction. Compared to the highly optimized industrial baseline VICC, Trailblazer achieves a 0.92\% relative reduction in overall video stall rates. Furthermore, it consistently outperforms VICC in terms of fine-grained stall metrics,  reducing moderate and severe stall rates by 1.28\% and 0.76\%, respectively. These performance advantages underscore the effectiveness of Trailblazer in mitigating generic, moderate and prolonged playback interruptions, which are disruptive to user experience. Given the massive scale of Douyin, which serves tens of millions of daily active users, even marginal gains in stall rates can yield substantial real-world benefits. To evaluate how these gains can translate into platform-wide industrial impacts, we establish a formal projection framework as illustrated in Fig.~\ref{fig:online_test}C. Based on this mathematical model, Fig.~\ref{fig:online_test}B illustrates the projected daily reduction in video stall time. Specifically, Trailblazer achieves a significant daily reduction in video stall time, cutting generic video stall time by 3,145.41\,hours, moderate video stall time by 588.21\,hours, and severe video stall time by 197.37\,hours per day. Collectively, these results provide strong empirical validation:  empowered by Trailblazer, the LLM can not only operate reliably in production network environments but also introduce industrial improvements in service quality.

While the previous results show the overall benefits, we next investigate the generalization capabilities of  Trailblazer across real-world diversity. First, we delve into the study of its performance across heterogeneous operating systems (OS) reflecting the various hardware and software configurations (e.g., Apple-series devices, application versions). Fig.~\ref{fig:online_test}D shows the distribution of different OS categories and compares the stall rates between Trailblazer and VICC across these categories. Results indicate that Trailblazer achieves performance on par with VICC on Android, while reducing video stall rates by 3.58\% on iOS. In particular, it  significantly outperforms VICC by 24.45\% on other less common OS platforms which mainly consist of latest versions of HarmonyOS. This significant gain could be attributed to the fact that VICC, being a rule-based policy, has not yet been fully optimized for these emerging platforms. In contrast, by leveraging the LLM as the foundation, Trailblazer enables rapid OS adaptation without manual tuning, and thus exhibits strong generalization across heterogeneous and evolving device systems. The results demonstrate the suitability of Trailblazer for LLM deployment in real-world scenarios where users span a wide range of devices and OS platforms. Finally, we investigate the impact of geographical distance on  the performance of Trailblazer by  categorizing clients based on their physical distance from the server. The geographical distance serves as a proxy for key network characteristics that affect video streaming quality, such as inter-domain routing complexity and transmission path stability. The distribution of clients across different regions and the corresponding performance results are presented in Fig.~\ref{fig:online_test}E. As shown in Fig.~\ref{fig:online_test}E, Trailblazer consistently outperforms VICC in reducing video stall rates across all regions, with the relative reduction of 0.86\%–10.29\%. Notably, the performance gain increases with distance, indicating that Trailblazer is particularly effective in complex network environments. Hence, Trailblazer demonstrates strong generalization and robustness to network variability, enabling large-scale deployment without region-specific customization.

In summary, the large-scale online evaluation on Douyin provides strong empirical evidence for the generalization capabilities of LLMs in networking. Furthermore, these results confirm the effectiveness of the proposed Trailblazer in adapting LLMs  and addressing the efficiency bottlenecks of LLM-based network control. Enabled by Trailblazer, the LLM not only performs reliably within real-world, concurrent and latency-sensitive network systems, but also exhibits robust generalization across diverse settings without requiring manual and scenario-specific tuning. 

\section*{Discussion}

For decades, the design of network policies has been dominated by handcrafted control rules or specialized deep learning models. However, these approaches often encounter a generalization bottleneck and require labor-intensive manual updates to accommodate the dynamic evolution of network environments, leading to significant engineering overhead. While LLMs offer a transformative opportunity for generalizable network control, their practical adoption in networking remains hindered by the misalignment between NLP and networking, as well as the computational inefficiency arising from their large model scale. To bridge this gap, we propose Trailblazer, a systematic framework that incorporates a domain alignment scheme to adapt LLMs for networking and an adaptive policy collaboration mechanism to reduce unnecessary LLM computation via intelligent request routing. Extensive experiments demonstrate that the Trailblazer-adapted LLM significantly outperforms conventional policies with superior generalization  across heterogeneous tasks and network environments. Furthermore, a three-week large-scale online A/B test on Douyin confirms that, enabled by Trailblazer, the LLM not only surpasses the highly optimized industrial baseline VICC in key metrics but also remains reliable in production environments. It delivers  service improvements across diverse settings without any scenario-specific manual tuning. This provides evidence that the generalization capabilities of LLM-based policies can translate into measurable benefits in real-world control settings, where operating conditions cannot be exhaustively captured during training.
Beyond the empirical success of LLMs for network control, we further uncover two counterintuitive yet critical insights. The first is early saturation, which reveals that  the standard scaling laws of LLMs do not strictly hold in networking tasks. Instead, with proper domain alignment, even small-scale LLMs can exhibit sufficient generalization. The second insight refers to selective invocation. In most common scenarios, lightweight conventional policies provide satisfactory performance. Designating LLMs as intelligent experts invoked only under complex network conditions achieves a better balance between efficiency and performance. Together, these insights provide solid support for the design of Trailblazer--the alignment scheme enables generalization for small-scale LLMs, and the collaboration mechanism allows LLM-based policies to efficiently operate within real-time network systems. 

Trailblazer also resonates with the emerging view of harness engineering~\cite{lin2026agentic} for foundation models. In  programming tasks, harnesses mainly focus on improving agent productivity by providing tools, execution environments, and feedback. Trailblazer extends this concept from productivity-oriented tasks to latency-sensitive decision  systems, where harnesses must determine how real-world states are represented, how valid actions are generated, how domain-specific control knowledge is acquired, and when model inference should enter the control loop. From this perspective, Trailblazer offers a general framework for designing decision-system harnesses through state-action alignment, task-level adaptation, and efficient policy collaboration.

In summary, our work demonstrates the potential of LLMs to serve as the foundation for generalizable network policies and represents an important step toward a practical paradigm for designing future network policies with stronger generalization and reduced manual intervention. More broadly, our findings suggest that effective foundation-model-based control relies more on harnesses that ground models as control policies and integrate them into the control loop than on model scaling or extensive invocation. Furthermore, our work provides a practical framework for adapting and deploying foundation models in other scientific and engineering systems, such as chemical process control, embodied robotic manipulation, and autonomous energy management, where generalization and operational efficiency are both essential.

\clearpage 

%
\bibliography{main} 
\bibliographystyle{sciencemag}

%
%
%
%
%
%


\section*{Acknowledgments}

\paragraph*{Author contributions:}
D.W. led the project, designed the overall framework, implemented the core methods, conducted all experiments (with others below) and took the lead in writing the manuscript. L.J.K. helped to write the original draft, revise the manuscript, perform the CC experiments, and analyze the experiment results. Z.M.W. helped to revise the manuscript, perform the real-world deployment experiments, and analyze the experiment results. F.X.W. partially supervised the project, and provided feedback and funding. W.Z. supervised the CC experiments, and contributed to the real-world deployment experiments. C.B.S. helped to review the manuscript, provide critical feedback on the manuscript and refine the presentation of the work. X.F.T. helped to perform the real-world deployment experiments. W.Y. and L.Z. provided the resources for real-world deployment experiments. W.W.Z. and P.C. helped to review the manuscript and provide important feedback. Z.W. initiated and supervised the project, guided the overall research vision, provided funding and contributed to manuscript reviewing.
\paragraph*{Competing interests:} 
There are no competing interests to declare.

\paragraph*{Data and materials availability:}
The datasets can be accessed through our open-source GitHub repository: \url{https://github.com/duowuyms/Trailblazer}. 
The data supporting the study of industrial deployment are available from Douyin, but  are not publicly available due to licensing restrictions. However, data may be obtained from the authors upon reasonable request and with the permission of Douyin. Besides, we have released our code showing our implementation, including dataset processing, Trailblazer implementation, training and inference recipes, and simulator integration. The code is  available at:  \url{https://github.com/duowuyms/Trailblazer}.


\subsection*{Supplementary materials}
Methods\\
Supplementary Text\\
Figs. S1 to S3\\
Tables S1 to S3


\newpage


\renewcommand{\thefigure}{S\arabic{figure}}
\renewcommand{\thetable}{S\arabic{table}}
\renewcommand{\theequation}{S\arabic{equation}}
\renewcommand{\thepage}{S\arabic{page}}
\setcounter{figure}{0}
\setcounter{table}{0}
\setcounter{equation}{0}
\setcounter{page}{1} 


\begin{center}
\section*{Supplementary Materials for\\ \scititle}


\author{
	Duo Wu,
	Linjia Kang,
	Zhimin Wang,
    Fangxin Wang,\\
    Wei Zhang,
    Chongbo Sun,
    Xuefeng Tao,
    Wei Yang,\\
    Le Zhang,
    Wenwu Zhu,
    Peng Cui,
    Zhi Wang$^{\ast}$\\
	\small$^\ast$Corresponding author. Email: wangzhi@sz.tsinghua.edu.cn
}
\end{center}

\subsubsection*{This PDF file includes:}
Methods\\
Supplementary Text\\
Figures S1 to S3\\
Tables S1 to S3


\newpage


\section{Methods}
\label{sec:methods}
\subsection*{Trailblazer implementation}
This section describes the implementation of Trailblazer across three networking tasks: ABR, CJS, and CC. Specifically, we detail the design of domain alignment scheme in each task, including network state encoder, network action decoder, and offline reinforcement fine-tuning algorithm. We further present the design of the adaptive policy collaboration in the CC task, including the request scheduler that enables efficient collaboration, as well as specifications of the lightweight CC policy used to collaborate with the LLM. 

\noindent\textbf{Network state encoder.} The network state encoder projects multimodal network information into token-like embeddings, which relies on the feature encoders to effectively extract features from modality-specific data. Instead of handcrafting specialized feature encoders from scratch, which will entail prohibitive model engineering costs, we employ the well-established model architectures tailored to each input modality (summarized in Table S1). Specifically, we leverage the graph neural network (GNN)~\cite{wu2020comprehensive} to encode the DAG data in CJS, 1D convolutional neural network (CNN)~\cite{li2022convolutional} to encode vectorized sequential data such as historical throughput in ABR, and simple linear layer to encode scalar data such as buffer occupancy in ABR and the loss rate in CC.

\noindent\textbf{Network action decoder.} The network action decoder is designed as a linear layer to generate task-specific control actions based on the LLM output features, which can be flexibly customized according to the action specifications of different tasks (outlined in Table S1). To be more specific, for ABR, we implement the network action decoder as a linear layer that computes a probability distribution over candidate bitrates, with the highest-probability bitrate selected for video download. For CJS, we use two parallel action decoders: one predicts the next stage to execute, and the other determines the number of executors to allocate. For CC, a linear regression layer is used as the action decoder for continuous bandwidth prediction.

\noindent\textbf{Offline reinforcement fine-tuning algorithm.}
We design the offline reinforcement fine-tuning algorithm based on the decision transformer (DT)~\cite{chen2021decision} because of its efficiency. Formally, DT reformulates RL as a sequence modeling problem, which seamlessly caters to the sequence modeling nature of the LLM. 
To be specific, we first employ the non-LLM policies for environment interaction and collect an experience dataset which consists of diverse experience trajectories: $\mathcal{D}_{DT}=\{\tau_1, \cdots, \tau_{|\mathcal{D}_{DT}|}\}$. Each trajectory $\tau = \{r_i, s_i, a_i\}_{i=1}^T$ is composed of rewards $r$, states $s$ and actions $a$, where $T$ denotes the episode length. Next, we fine-tune the LLM over this experience dataset, where  the LLM takes the historical returns, network states, and network actions to predict the next action: 
\begin{equation}
    LLM(\hat{a}_i\mid R_{i-w}, s_{i-w}, a_{i-w}, \cdots, R_i,s_i).
\end{equation}
Here, the return $R_i=\sum_{t=i}^T r_t$ denotes the cumulative rewards expected to receive from state $s_i$, and $w$ denotes the historical context window for effective learning. The training loss is designed to minimize the discrepancy between predicted action $\hat{a}_i$ and ground-truth action $a_i$: 
\begin{equation} \mathcal{L}_i = {F}(\hat{a}_i, a_i), \end{equation} 
where $F$ can be cross-entropy loss for discrete actions or mean squared error for continuous actions. The core idea of this algorithm is to train the LLM to learn the distribution of actions conditioned on specific states and returns, enabling the LLM to generate actions to achieve the desired return after training. In particular, during inference, we specify a sufficiently high target return indicating the desired performance to trigger the LLM to generate high-quality actions.

In addition to the DT-based approach, we also propose another variant, the contextual imitation learning (CIL) algorithm, which leverages  near-optimal expert demonstrations to guide the LLM toward an optimal  policy. Specifically, each trajectory in the experience dataset $\mathcal{D}_{CIL}=\{\tau_1, \cdots, \tau_{|\mathcal{D}_{CIL}|}\}$ is augmented with expert actions $a^e$, such that $\tau = \{r_i, s_i, a_i, a_i^e\}_{i=1}^T$. The LLM predicts the next action $\hat{a}_i$ conditioned on a context window of historical states and actions:
\begin{equation}
    LLM(\hat{a}_i\mid s_{i-w},a_{i-w},\cdots, s_{i-1}, a_{i-1},s_i),
\end{equation}
where $w$ denotes the context length. The model is trained by minimizing the difference between the predicted and expert actions:
\begin{equation}
\mathcal{L}_i=F(\hat{a}_{i}, a^e_i),
\end{equation}
Note that unlike conventional imitation learning~\cite{zheng2022imitation,zare2024survey} that typically relies on short-term observations, CIL explicitly maintains a long context window to exploit the LLM’s powerful capability in capturing temporal patterns and contextual dependencies in network dynamics, thereby enabling more effective policy learning.

The choice between DT and CIL to implement our offline reinforcement fine-tuning algorithm hinges on the availability of near-optimal expert demonstrations during training. When such high-quality supervision signals are accessible, CIL is preferred: it directly learns to imitate optimal behavior without requiring return estimation, simplifying the learning objective. Moreover, by removing the need to encode the return information, CIL reduces input sequence complexity and context length, leading to faster inference and lower computational overhead. In contrast, DT, which conditions action prediction on returns, becomes more suitable in settings where near-optimal actions are not available, but historical trajectories with associated returns can be reconstructed. In such cases, the return serves as a behavioral prior that guides action selection toward high-reward outcomes, even in the absence of explicit expert demonstrations. Thus, CIL and DT represent complementary paradigms: CIL excels in simulation-based offline training with oracle supervision, while DT is better suited for reward-driven learning from suboptimal traces.

In practical implementation, we use the DT-based algorithm for ABR and CJS tasks, where near-optimal actions are difficult to obtain. The reward in ABR is defined as the QoE score, while the reward in CJS is represented as the negative JCT. The cross-entropy loss is selected as the loss function since these tasks involve discrete control actions. Besides, the context window $w$ is empirically set to 10 and 20 for ABR and CJS, respectively. For the CC task, we adopt the CIL-based algorithm for fine-tuning, where we leverage ground-truth bottleneck bandwidths--obtained from network simulations--as near-optimal expert demonstrations to guide the LLM toward optimal congestion control policies. Although the true bottleneck bandwidth is unobservable in real-world network systems, it can be accurately inferred in controlled simulation environments during training, serving as a high-quality approximation of the optimal sending rate that maximizes bandwidth utilization. We therefore use the simulated bottleneck bandwidth as the expert action to supervise the LLM. The loss function is implemented as mean squared error for continuous bandwidth prediction and the context window $w$ is set to 10.

\noindent\textbf{Adaptive policy collaboration.} Due to the high inference latency of the LLM, it is impractical to deploy the fine-tuned LLM for fine-grained per-flow control, especially in the CC service of Douyin, where requests must be processed within 100\,ms under high concurrency. Hence, we activate the  adaptive policy collaboration mechanism  of Trailblazer to improve the efficiency of LLM-based network systems while maintaining high performance. Notably, we design the scheduler as a lightweight rule-based classifier to distinguish between flows under good and poor network conditions. While more sophisticated schedulers (e.g., learning-based) are possible, we employ a rule-based design to ensure minimal overhead and fast processing speed, which is critical for routing the large volume of concurrent requests in real-world network systems. Specifically, the request of a flow is classified as experiencing \textit{good network conditions} if: (i) its last round-trip time is below a threshold $\alpha_1$; (ii) the packet loss rate is below $\alpha_2$; and (iii) the last sending rate exceeds $\alpha_3 \times {rate}^{{req}}$. Flow requests meeting all three criteria are routed to a lightweight policy while all others are directed to the LLM for intelligent control. Here, ${rate}^{{req}}$ is the request rate, which denotes the application-specific upper bound on sending rate required to maintain the quality of video transmission. The request rate is computed by a dedicated module in Douyin. In practice, the three thresholds are set to $\alpha_1=50$ (ms), $\alpha_2=0.05, \alpha_3=0.95$. 

The lightweight policy used to collaborate with the LLM is simply  designed to set the sending rate directly to ${rate}_{{req}}$. While this strategy tends to exhaust available bandwidth and induce congestion under poor network conditions, it performs reliably in general cases, where network conditions are stable and ${rate}_{{req}}$ is often lower than actual bottleneck link capacity. This collaboration  thus enables scalable deployment: the LLM is invoked only when necessary, while the majority of flow requests are handled efficiently by the lightweight policy. To further enhance system efficiency and reduce per-request processing latency, we use the compact Qwen2.5-0.5B~\cite{qwen2025qwen25technicalreport} as the foundation LLM and set the batch size to 64, achieving an average inference latency of 37.1\,ms. This meets Douyin’s 100\,ms response latency requirement for the CC task while preserving a safety margin to support system resilience.  Fig.~\ref{fig:figures3} offers a detailed analysis of the impact of LLM inference batch size on CC task performance.

\subsection*{Experimental setup details}\label{sec:exp_details}


\subsubsection*{Experimental details of ABR}

\noindent\textbf{Simulator.} We employ the widely adopted ABR simulator developed by Mao et al.~\cite{mao2017neural}, which accurately models the dynamics of real-world video streaming applications through stateful representations of playback buffer and network environment. Given a bandwidth trace and video manifest file, the simulator computes the download time for each video chunk based on its bitrate and the current available bandwidth. It then drains the buffer by the download time to reflect ongoing playback, and adds the duration of the downloaded chunk to the buffer to simulate buffer occupancy. Owing to the faithful simulation of video streaming process, this simulator provides a flexible and efficient approach to evaluate ABR policies in diverse environments by varying the input bandwidth traces and video manifest files.

\noindent\textbf{Dataset.} To evaluate the generalization capability of each method under diverse and challenging network environments, we employ a combination of real-world and synthetic datasets. For network dynamics, we use the \textit{FCC}~\cite{fcc} broadband measurement dataset as the primary source of bandwidth traces, which captures real-world variability across a representative sample of USA consumers. From this collection, we randomly select 485 traces, with 235 for training, 150 for validation, and 100 for testing. This results in the use of more than 324,000 seconds of bandwidth traces for experiments. Additionally, to evaluate generalization under more challenging network dynamics, we synthesize a more dynamic bandwidth dataset \textit{SynthTrace} that comprises 100 traces with broader bandwidth ranges and more dynamic fluctuation patterns. For video content, we adopt the widely used \textit{Envivio-Dash3}~\cite{dashvideo} as the default video, and further introduce a synthetic video \textit{SynthVideo} with larger chunk sizes for performance evaluation. Based on these datasets, we simulate various environments for generalization evaluation, as presented in  Table~\ref{table:abr_settings}.

\noindent\textbf{Baseline.} We compare Trailblazer with state-of-the-art ABR policies: the learning-based method GENET~\cite{xia2022genet}, and rule-based methods, BBA~\cite{huang2014bba} and MPC~\cite{yin2015mpc}. GENET leverages RL  to train a small neural model to optimize video streaming and improve the training process with  curriculum learning. BBA heuristically considers buffer occupancy as a critical signal for bitrate control and designs an algorithm to maintain the playback buffer occupancy at a desired level. MPC leverages both throughput estimates and buffer occupancy to choose bitrates by optimizing a given QoE metric over a future time window.

\noindent\textbf{Experience dataset construction.} To construct the experience dataset for fine-tuning the LLM with Trailblazer, the baseline GENET is adopted for experience collection. Specifically, we simulate diverse network environments, use GENET to interact with these environments, and collect  interactions in terms of states, actions and rewards as the training dataset for LLM fine-tuning.

\noindent\textbf{Metric.} Following the common practice in ABR~\cite{mao2017neural,yin2015mpc,xia2022genet}, we adopt QoE as the primary performance metric, which is defined as the weighted linear combination of three metrics: 
\begin{equation}
    {QoE}_i = {bitrate}_i - \lambda_1 \times {rebuf}_i - \lambda_2 \times |\Delta{bitrate}_i|, 
\end{equation}
where ${bitrate}_i$ is the bitrate (in Mbps) of chunk $i$, ${rebuf}_i$ is the rebuffering time (in seconds) during its download, and $|\Delta{bitrate}_i| = |{bitrate}_i - {bitrate}_{i-1}|$ is the absolute bitrate change between consecutive chunks. The coefficients $\lambda_1$ and $\lambda_2$ control the trade-offs among these factors. Based on prior work~\cite{mao2017neural}, we set $\lambda_1 = 4.3$ and $\lambda_2 = 1$, reflecting the higher user sensitivity to rebuffering compared to bitrate fluctuations.

\subsubsection*{Experimental details of CJS}

\noindent\textbf{Simulator.} We use the open-source Apache Spark job scheduling simulator~\cite{spark_sim} to model task execution, resource allocation, and scheduling logic in Spark clusters. Given a workload trace consisting of a sequence of jobs and a number of executors representing the cluster computing resources, the simulator schedules tasks according to the specific policy within an emulated cluster environment. This enables controlled evaluation of scheduling policies under conditions that approximate real-world workloads.

\noindent\textbf{Dataset.} The TPC-H benchmark~\cite{tpch} is used as the primary source of workload traces, which consists of a suite of business-oriented computational jobs that have broad industry-wide relevance. These jobs are characterized by large data volumes and high processing complexity, making TPC-H a widely adopted benchmark for job scheduling. By randomly sampling jobs from the TPC-H suite and varying the number of executor resources in the simulated cluster, we generate diverse experimental environments that span different workload intensities and resource availability. The simulation configurations are summarized in Table~\ref{table:cjs_settings}.

\noindent\textbf{Baseline.} Similar to ABR, we implement one simple learning-based policy Decima~\cite{mao2019cjs} and two rule-based policies  FIFO~\cite{spark_scheduler} and Fair~\cite{spark_scheduler} for performance comparison. Decima is an RL model for job scheduling with a GNN model to process job DAG information. Both FIFO and Fair are common scheduling policies used by big data processing system Spark~\cite{spark_scheduler}. The former schedules jobs in the order of their arrival and allocates the requested amount of resources to each job, while the latter schedules jobs in a ``round robin'' fashion to ensure that each job receives a roughly equal share of the cluster resources.

\noindent\textbf{Experience dataset construction.} The experience dataset is constructed by using the Decima model to interact with diverse simulated training environments. The resulting state-action-reward tuples are used as the training fuel to fine-tune the LLM based on Trailblazer.

\noindent\textbf{Metric.} We use JCT as the evaluation metric following the literature~\cite{mao2019cjs}. Let $t_s$ denote the job arrival time and $t_e$ denote the job finishing time. The JCT is calculated by:
\begin{equation}
    JCT=t_e-t_s.
\end{equation}

\subsubsection*{Experimental details of CC}
\noindent\textbf{Baseline.} The production-grade VICC policy is used as the primary baseline across all experiments. Deployed in Douyin's real-time services, VICC is an adaptive policy designed to balance bandwidth utilization and latency across global network conditions and diverse audiovisual applications. It has been iteratively refined over years and incorporates multiple adaptive mechanisms such as congestion response, bandwidth probing, packet loss detection, and jitter resilience. These capabilities enable dynamic adaptation to changing network states and application priorities, achieving high bandwidth efficiency while meeting strict application-specific latency requirements. As a result, VICC enhances the quality of real-time media transmission in practice, serving as a strong, production-grade baseline in our evaluation.

\noindent\textbf{Experience dataset construction.} Based on Douyin's internal development platform, we construct the experience dataset to fine-tune the LLM for the CC task. Specifically, we establish real-world video sessions using six mobile devices and nine types of media content (e.g., singing and gaming), and impose diverse and complex network conditions (e.g., varying bandwidth, delay, and packet loss) using the enterprise-grade network emulator HoloWAN~\cite{holowan}. The device, media, and network configurations are  aligned with Douyin's online settings. During data collection, for each session, CC decisions are determined by one of the four rule-based policies randomly selected from Douyin. This process yields an experience dataset comprising over 30,000 sessions and more than 10 million data samples. The dataset is further split into a 95\% training subset for LLM fine-tuning and a 5\% test set for performance validation, including model scale investigation and evaluation of the selective invocation mechanism.

\noindent\textbf{Online deployment.}
For online A/B tests, we deploy six media servers that are responsible for ensuring service quality for media sessions and implement the CC logic. Incoming flow requests are randomly routed to these servers. Among them, three run the VICC policy, while the other three are configured with Trailblazer for intelligent control. In Trailblazer, the scheduler  and the lightweight CC policy are deployed locally on the media servers. The LLM, in contrast, is hosted on a commercial GPU server. {We use Qwen2.5-0.5B as the LLM and set the inference batch size to 64. With these configurations, the LLM consumes approximately 4.5\,GB GPU memory and takes about 30\,ms per inference.} The scheduler and the LLM communicate via the WebSocket API. When the scheduler identifies incoming flows experiencing poor network conditions, it forwards their contextual information to the LLM. Upon completion of inference, the LLM returns control decisions to the scheduler for execution.  {By combining a small-scale LLM, batched inference, and collaborative control, Trailblazer can handle the concurrent requests assigned by Douyin to the three media servers. It consistently keeps end-to-end request processing latency  below 100\,ms, which satisfies the real-time requirements of CC in Douyin.} 


\noindent\textbf{Metric.}
For offline performance validation of Trailblazer, we adopt MAPE as the primary metric, which quantifies the relative discrepancy between the predicted bandwidth and bottleneck link capacity or request rate. Given the predicted bandwidth $b^p_i$, the true simulated bottleneck bandwidth $b^t_i$ and the request rate $rate^{req}_i$, MAPE is computed by:
\begin{equation}
{MAPE} = \frac{1}{n} \sum_{i=1}^n \left| \frac{\min (b^p_i, rate^{req}_i) - \min(b^t_i, rate^{req}_i)}{\min(b^t_i, rate^{req}_i)} \right|,
\end{equation}
where $n$ is the number of samples in the validation dataset.

To evaluate system efficiency, we measure request processing delay, which is defined as the time interval between the arrival of a flow request at the system and the receipt of the corresponding CC decision. This delay consists of LLM queuing delay and inference latency. Under high request load, queuing delay becomes dominant because LLM inference latency remains relatively stable for a fixed processing batch size.

For online A/B tests, video stall rate is used as the primary performance metric, which is a  key industrial indicator that directly reflects playback smoothness and user experience. The stall rate is calculated as the ratio of the total stall duration to the total playback duration across all flows:
\begin{equation}
{stall\_rate} = \frac{\sum_{i=1}^N d_i^s}{\sum_{i=1}^N d_i^p},
\end{equation}
where $d_i^s$ denotes the cumulative stall duration of flow $i$ as continuously reported by a monitor, $d_i^p$ denotes the total playback duration of flow $i$, and $N$ is the total number of flows. A lower stall rate generally contributes to improved user retention and engagement, making it highly relevant to business outcomes. The relative reduction of Trailblazer over VICC is then computed by:
\begin{equation}
    reduction=\frac{stall\_rate_{VICC}-stall\_rate_{Trailblazer}}{stall\_rate_{VICC}}.
\end{equation}


\section*{Supplementary Text}


{\noindent\textbf{LLM usage compliance.} All LLMs used in the experiments comply with their respective license terms. Simulation-based experiments  were carried out at the university using open-source LLMs that permit academic research, including Llama2. All online A/B tests  were conducted at ByteDance (Douyin) using Qwen2.5, an open-source LLM licensed for commercial use. Llama2 or other models restricted for commercial entities were not used in any stage of the industrial deployment or internal evaluation at ByteDance.}

\noindent\textbf{Data privacy.} The data collection and analysis in the large-scale online A/B tests comply with the agreement established between Douyin and its users. No personally identifiable information (e.g., phone number) was collected, and  the collected data cannot be linked to  users’ real identities. 
In estimating client-to-server distance, we define the distance as the great-circle distance between the client's city and the server's city. Hence, only city-level information was used for distance estimation, without tracking or inferring users' exact locations. The analysis was conducted within Douyin’s internal secure platform which avoids data leakage.








\begin{figure}
    \centering
    \includegraphics[width=0.995\linewidth]{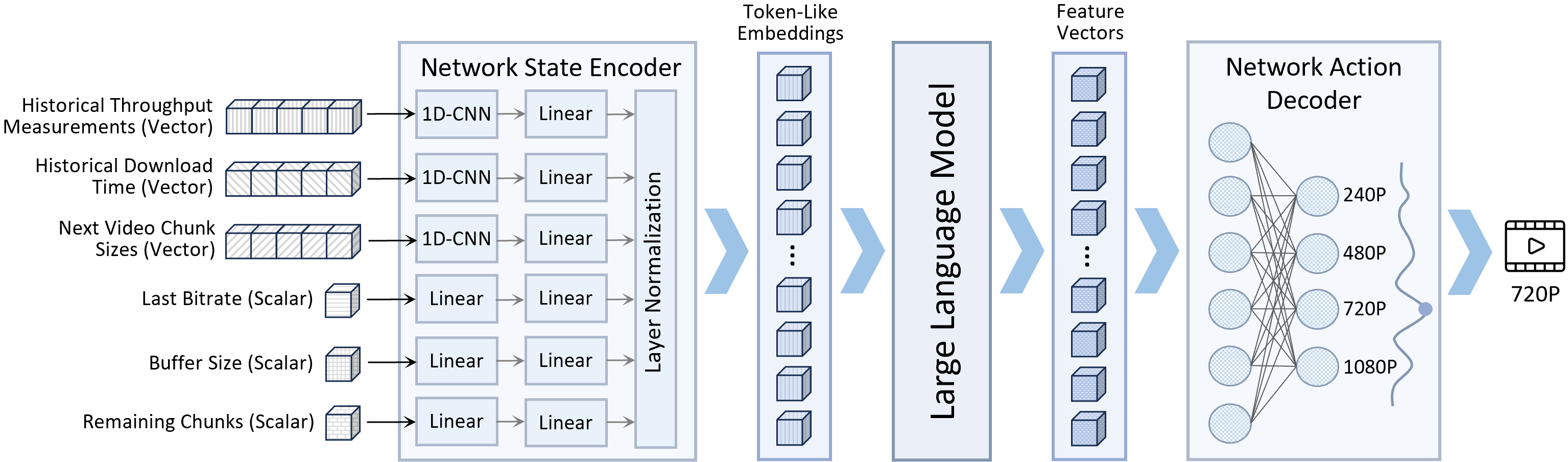}
    \caption{\textbf{Illustration of the design of network state encoder and action decoder with ABR task as the example.} The network state encoder employs 1D CNNs to extract structural features from vectorized information, while using linear layers to process scalar data.  These diverse features are independently projected into a semantic space aligned with language tokens and subsequently fused via layer normalization.  The network action decoder is implemented as a linear layer that maps the high-dimensional feature vectors produced by the LLM into a probability distribution over candidate bitrate versions.  The bitrate with the highest probability is selected as the action for downloading the next video chunk. Through the above adaptation, the LLM can process state information and generate control actions for the ABR task.}
    \label{fig:figures1}
\end{figure}

\clearpage

\begin{figure}
    \centering
    \includegraphics[width=0.98\linewidth]{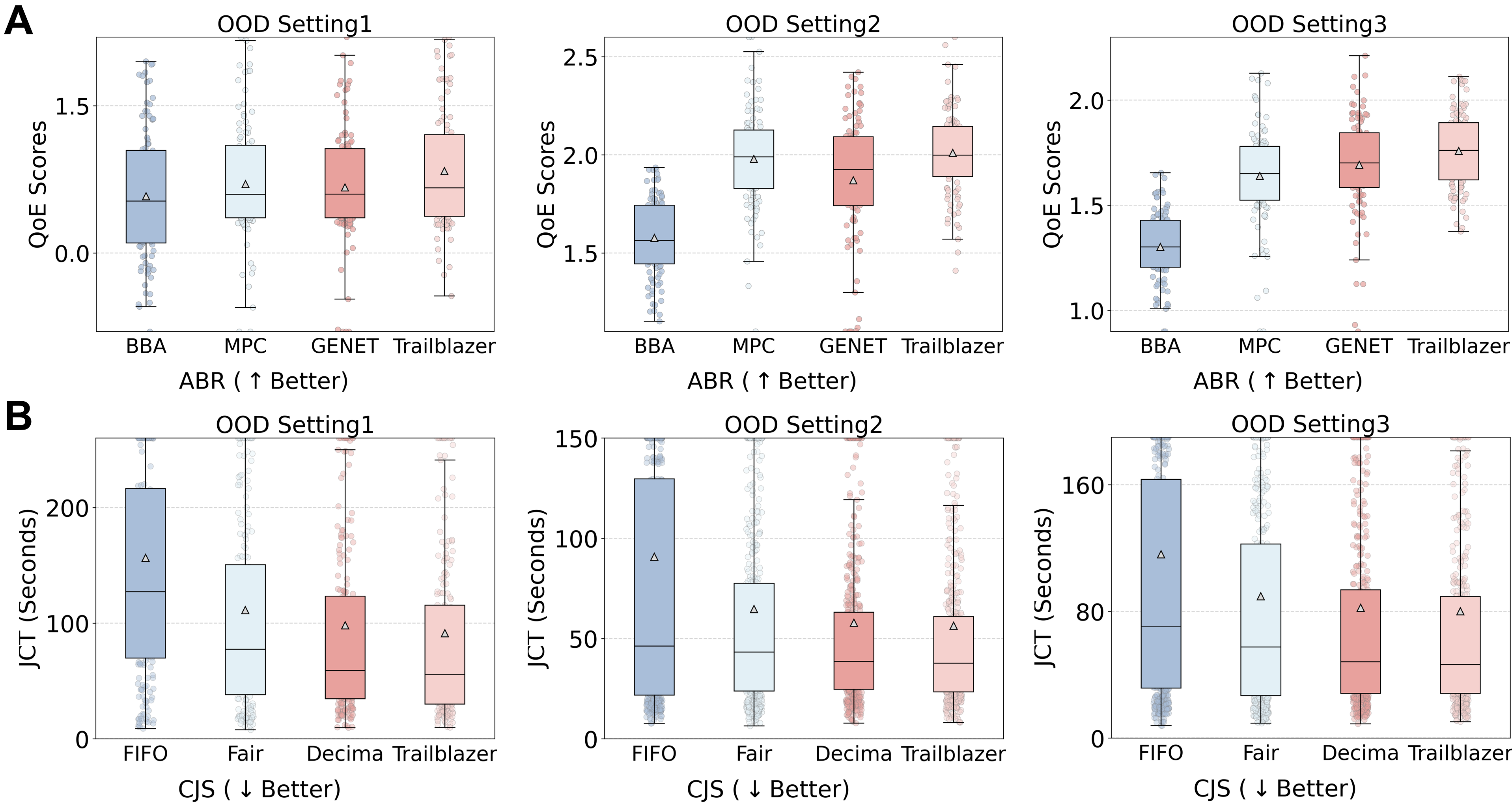}
    \caption{\textbf{Performance comparison of generalization under various OOD test settings. } Scatters and box shapes represent the distribution of performance, while triangles denote mean values.
    (\textbf{A})
    Results in the ABR task across three OOD settings. 
    (\textbf{B})
    Results in the CJS task across three OOD settings.
}
    \label{fig:figures2}
\end{figure}

\clearpage

\begin{figure}
    \centering
    \includegraphics[width=0.98\linewidth]{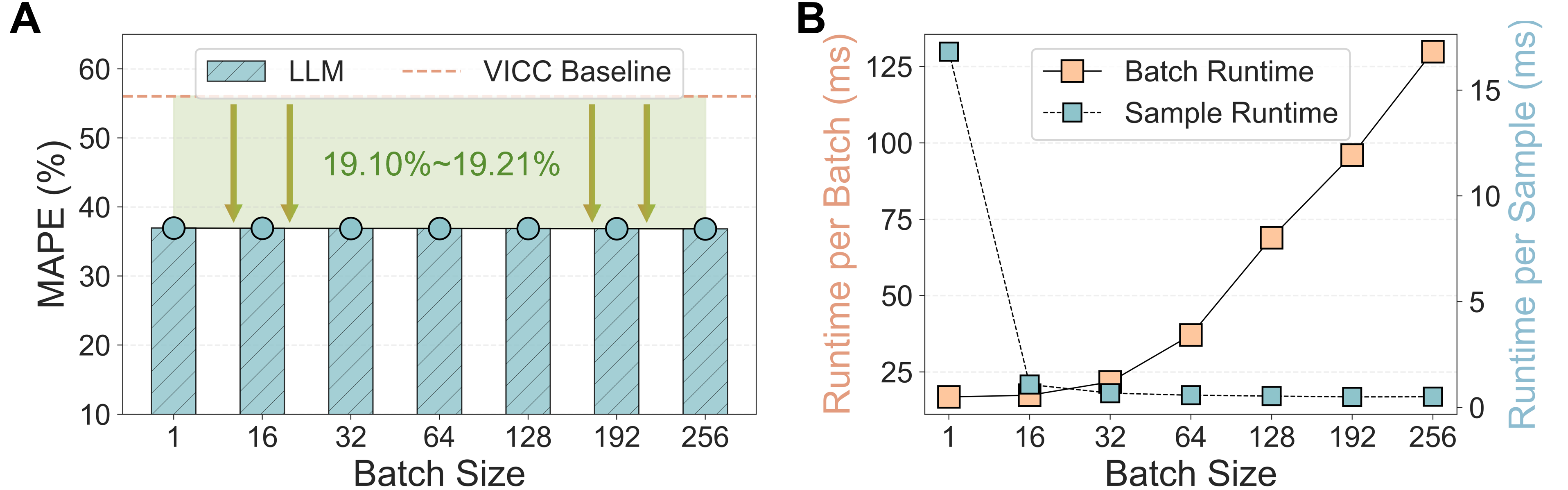}
    \caption{\textbf{Study of the impact of LLM inference batch size on the CC task. } We use the Qwen2.5-0.5B as the backbone LLM for this study.
    (\textbf{A})
    Performance comparison across different batch sizes. The dashed line indicates the performance of VICC.
    (\textbf{B})
    Inference runtime analysis under different batch sizes. Increasing the LLM batch size can effectively reduce the per-request inference latency without compromising task performance. 
    Hence, we set the batch size to 64, achieving the average inference latency of 37.1\,ms, which satisfies 100\,ms response latency requirement of CC in Douyin while leaving a safety margin for system resilience.
}
    \label{fig:figures3}
\end{figure}

\clearpage


\begin{table}
    \centering
    \caption{\textbf{Overview of the three evaluation networking tasks.}}
    \label{table:table_s1}
    \renewcommand{\tabularxcolumn}[1]{>{\centering\arraybackslash}m{#1}} 
    \begin{tabularx}{\textwidth}{
        >{\centering\arraybackslash}w{c}{2cm}   
        X                                        
        X                                        
        X                                        
    }
        \hline
        \textbf{Task} & \textbf{State Modalities} & \textbf{Action Specifications} & \textbf{Objective} \\
        \hline
        ABR & \textit{vector}: historical throughputs; \textit{scalar}: buffer occupancy 
            & bitrate for the next video chunk 
            & maximize user's QoE \\
        CJS & \textit{graph}: DAGs describing dependency and resource demands of job execution stages 
            & job stage to run next, number of executors allocated to the stage 
            & minimize JCT \\
        CC  & \textit{scalar}: loss rate, delay 
            & bottleneck link bandwidth estimation 
            & maximize bandwidth utilization \\
        \hline
    \end{tabularx}
\end{table}

\clearpage

\begin{table}
    \centering
    \caption{\textbf{Summary of the environment settings for generalization evaluation in ABR.} }
    \label{table:abr_settings}
    \begin{tabularx}{\textwidth}{>{\centering\arraybackslash}X 
                                  >{\centering\arraybackslash}X 
                                  >{\centering\arraybackslash}X}
        \hline
        \textbf{Setting} & \textbf{Video} & \textbf{Bandwidth Traces} \\
        \hline
        {Training Setting} & \textit{Envivio-Dash3} & \textit{FCC} \\
        {ID Test Setting} & \textit{Envivio-Dash3} & \textit{FCC} \\
        {OOD Setting 1} & \textit{Envivio-Dash3} & \textit{SynthTrace} \\
        {OOD Setting 2} & \textit{SynthVideo} & \textit{FCC} \\
        {OOD Setting 3} & \textit{SynthVideo} & \textit{SynthTrace} \\
        \hline
    \end{tabularx}
\end{table}

\clearpage

\begin{table}
    \centering
    \caption{\textbf{Summary of the environment settings for generalization evaluation in CJS.}}
    \label{table:cjs_settings}
    \begin{tabularx}{\textwidth}{>{\centering\arraybackslash}X 
                                  >{\centering\arraybackslash}X 
                                  >{\centering\arraybackslash}X}
        \hline
        \textbf{Setting} & \textbf{Number of Job Requests} & \textbf{Number of Executors (k)} \\
        \hline
        {Training Setting} & 200 & 50 \\
        {ID Test  Setting} & 200 & 50 \\
        {OOD Setting 1} & 200 & 30 \\
        {OOD Setting 2} & 450 & 50 \\
        {OOD Setting 3} & 450 & 30 \\
        \hline
    \end{tabularx}
    \vspace{-0.3cm}
\end{table}


\end{document}